\documentclass[conference]{IEEEtran}
\IEEEoverridecommandlockouts
\renewcommand{\baselinestretch}{0.93}
\usepackage{cite}
\usepackage{amsmath,amssymb,amsfonts}
\usepackage{graphicx}
\usepackage{textcomp}
\usepackage{algpseudocode}
\usepackage{comment}
\usepackage{algorithm}
\usepackage[table]{xcolor}
\usepackage{comment}
\usepackage{url}
\usepackage{hyperref}
\usepackage{eso-pic}
\usepackage{makecell}
\usepackage{subcaption}
\usepackage[compatibility=false]{caption}
\usepackage{booktabs}
\usepackage{tikz}
\usetikzlibrary{shapes.geometric, shapes.symbols, positioning}
\usepackage{pgfplots}
\pgfplotsset{compat=1.18}
\usepackage{listings}
\usepackage{xcolor}

\lstdefinelanguage{json}{
    basicstyle=\normalfont\ttfamily\footnotesize,
    commentstyle=\color{gray}\ttfamily,
    stringstyle=\color{red},
    numberstyle=\color{blue},
    showstringspaces=false,
    breaklines=true,
    frame=lines,
    backgroundcolor=\color{white},
    literate=
     *{0}{{{\color{blue}0}}}{1}
      {1}{{{\color{blue}1}}}{1}
      {2}{{{\color{blue}2}}}{1}
      {3}{{{\color{blue}3}}}{1}
      {4}{{{\color{blue}4}}}{1}
      {5}{{{\color{blue}5}}}{1}
      {6}{{{\color{blue}6}}}{1}
      {7}{{{\color{blue}7}}}{1}
      {8}{{{\color{blue}8}}}{1}
      {9}{{{\color{blue}9}}}{1}
      {:}{{{\color{black}{:}}}}{1}
      {,}{{{\color{black}{,}}}}{1}
      {\{}{{{\color{black}{\{}}}}{1}
      {\}}{{{\color{black}{\}}}}}{1}
      {[}{{{\color{black}{[}}}}{1}
      {]}{{{\color{black}{]}}}}{1},
}

\definecolor{background}{HTML}{FFFFFF}
\usepackage{float}
\usepackage{array}
\usepackage{balance}

\def\BibTeX{{\rm B\kern-.05em{\sc i\kern-.025em b}\kern-.08em
    T\kern-.1667em\lower.7ex\hbox{E}\kern-.125emX}}

\newcommand{\mycopyrightnotice}{\hfill\footnotesize }
\makeatletter
\def\ps@IEEEtitlepagestyle{%
  \def\@oddfoot{\mycopyrightnotice}%
  \def\@evenfoot{}%
}

\makeatother

\title{Dual-Modality IoT Framework for Integrated Access Control and Environmental Safety Monitoring with Real-Time Cloud Analytics}

\author{
\IEEEauthorblockN{
Abdul Hasib\IEEEauthorrefmark{1},
A. S. M. Ahsanul Sarkar Akib\IEEEauthorrefmark{2},
Nihal Das Ankur\IEEEauthorrefmark{3},
Anish Giri\IEEEauthorrefmark{4}
}
\IEEEauthorblockA{
\IEEEauthorrefmark{1}Department of IoT and Robotics Engineering,
University of Frontier Technology, Bangladesh\\
\IEEEauthorrefmark{2}Department of Robotics, Robo Tech Valley, Dhaka, Bangladesh\\
\IEEEauthorrefmark{3}Department of Electrical and Electronics Engineering,
Southeast University, Bangladesh\\
\IEEEauthorrefmark{4}Department of Computer Applications,
Bangalore University, Bangalore, India
}
\IEEEauthorblockA{
\IEEEauthorrefmark{1}sm.abdulhasib.bd@gmail.com,
\IEEEauthorrefmark{2}ahsanulakib@gmail.com,
\IEEEauthorrefmark{3}nihaldas007@gmail.com,
\IEEEauthorrefmark{4}giri.girianish@gmail.com
}
}

\begin{document}
\maketitle

\begin{abstract}
The integration of physical security systems with environmental safety monitoring represents a critical advancement in smart infrastructure management. Traditional approaches maintain these systems as independent silos, creating operational inefficiencies, delayed emergency responses, and increased management complexity. This paper presents a comprehensive dual-modality Internet of Things framework that seamlessly integrates RFID-based access control with multi-sensor environmental safety monitoring through a unified cloud architecture. The system comprises two coordinated subsystems: Subsystem 1 implements RFID authentication with servo-actuated gate control and real-time Google Sheets logging, while Subsystem 2 provides comprehensive safety monitoring incorporating flame detection, water flow measurement, LCD status display, and personnel identification. Both subsystems utilize ESP32 microcontrollers for edge processing and wireless connectivity. Experimental evaluation over 45 days demonstrates exceptional performance metrics: 99.2\% RFID authentication accuracy with 0.82-second average response time, 98.5\% flame detection reliability within 5-meter range, and 99.8\% cloud data logging success rate. The system maintains operational integrity during network disruptions through intelligent local caching mechanisms and achieves total implementation cost of 5,400 BDT (approximately \$48), representing an 82\% reduction compared to commercial integrated solutions. This research establishes a practical framework for synergistic security-safety integration, demonstrating that professional-grade performance can be achieved through careful architectural design and component optimization while maintaining exceptional cost-effectiveness and accessibility for diverse application scenarios.
\end{abstract}

\begin{IEEEkeywords}
Internet of Things, Access Control Systems, RFID Authentication, Environmental Safety Monitoring, Cloud Computing, ESP32, Real-time Systems, Edge Computing
\end{IEEEkeywords}

\section{Introduction}
Modern infrastructure management faces escalating demands for integrated solutions that unify physical security with environmental safety monitoring. Traditional approaches maintain these systems as independent silos, creating operational inefficiencies, delayed emergency responses, and increased management complexity. According to the International Facility Management Association, fragmented safety and security architectures contribute to 40\% longer emergency response times and increase operational costs by 25\% compared to integrated solutions \cite{ifma_report}. The proliferation of Internet of Things technologies offers unprecedented opportunities to bridge this gap through unified architectures that enable real-time data correlation, coordinated response mechanisms, and comprehensive analytics.

Current research in IoT-enabled access control predominantly focuses on isolated implementations with limited integration capabilities. Chen and Patel developed an RFID-based system achieving 97\% authentication accuracy but lacking cloud connectivity and historical analysis \cite{chen_access}. Similarly, safety monitoring systems typically operate independently; Wang et al. implemented flame detection with 96\% reliability but without personnel context integration \cite{wang_safety}. While cloud logging implementations exist \cite{rodriguez_cloud}, they often utilize proprietary platforms that limit accessibility, customization, and long-term sustainability. This fragmentation creates three critical limitations: absence of real-time cloud logging inhibits remote monitoring and historical analysis; separation of safety monitoring from access control prevents context-aware emergency responses; reliance on expensive proprietary platforms restricts adoption in resource-constrained environments.

These gaps are particularly significant considering economic realities. Commercial integrated systems typically cost 30,000+ BDT, placing them beyond reach for educational institutions, small businesses, and community facilities. Existing research either provides isolated functionality at lower costs or integrated capabilities at prohibitive prices, creating an accessibility gap that this research addresses through innovative architectural design and component optimization. The primary contributions of this work are:
\begin{enumerate}
    \item Development of a novel dual-modality IoT framework coordinating access control and safety monitoring through cloud synchronization while maintaining subsystem independence and fault tolerance
    \item Implementation of Google Sheets as an accessible cloud platform for real-time logging and historical analysis, eliminating recurring service fees and proprietary dependencies
    \item Design and validation of a multi-tier authentication system combining RFID verification with temporal constraints and local-cloud validation mechanisms
    \item Comprehensive experimental validation across diverse environments demonstrating professional-grade performance at 82\% reduced cost compared to commercial alternatives
    \item Open architecture design enabling customization and expansion for diverse application scenarios across educational, industrial, and residential domains
\end{enumerate}

\section{Literature Review}
\label{sec:literature}

RFID technology has evolved from basic identification to sophisticated authentication systems. Early implementations by Kumar et al. demonstrated 95\% accuracy using 125kHz passive tags with local validation \cite{kumar_rfid}. Subsequent work by Rodriguez and Garcia incorporated biometric verification, enhancing security at increased cost and complexity \cite{rodriguez_advanced}. Modern approaches focus on encryption; Patel et al. implemented AES-128 encryption on RFID transmission, improving security but increasing latency by 300ms \cite{patel_encryption}.

Safety monitoring has progressed from threshold-based detection to intelligent analysis. Traditional flame detection relied on ultraviolet sensors with fixed thresholds, exhibiting high false positive rates in variable conditions \cite{sharma_flame}. Recent approaches incorporate machine learning; Wang and Chen developed convolutional neural networks for visual flame detection achieving 97\% accuracy but requiring substantial computational resources \cite{wang_ml}. Water flow monitoring has similarly advanced; Garcia et al. implemented Hall-effect sensors with 98\% accuracy but without system integration \cite{garcia_flow}.

Cloud platforms for IoT have evolved from generic services to specialized frameworks. Initial offerings by Microsoft and Amazon provided comprehensive but complex solutions requiring significant expertise \cite{microsoft_iot,amazon_iot}. Simplified alternatives like ThingSpeak and Blynk offered easier implementation with reduced customization \cite{thingspeak,blynk}. Rodriguez et al. pioneered Google Sheets integration, demonstrating reliability with minimal configuration but focusing on single applications \cite{rodriguez_sheets}.

Limited research addresses integrated security-safety systems. International Building Automation proposed a BACnet-based framework requiring expensive proprietary hardware \cite{iba_integrated}. Technical University of Munich developed a theoretical model for security-safety convergence without practical implementation \cite{tum_integrated}. These gaps underscore the novelty of our approach in providing practical, affordable integration.

Recent advances in edge computing enable sophisticated processing at reduced costs. Akib et al. demonstrated effective edge AI deployment for real-time monitoring \cite{fall}, while their modular system principles inform our architectural design \cite{akib2}. The educational robotics platform by Giri et al. offers insights into cost optimization through hardware selection and software efficiency \cite{edubot}.

Table \ref{tab:literature_comparison} summarizes key characteristics of related systems, highlighting research gaps addressed by our framework.

\begin{table*}[ht]
\centering
\caption{Comparison of Related Systems and Research Gaps}
\label{tab:literature_comparison}
\begin{tabular}{@{}p{2.5cm}p{1.8cm}p{1.8cm}p{1.8cm}p{2.5cm}@{}}
\toprule
\textbf{System Type} & \textbf{Cloud Integration} & \textbf{Hardware Cost (BDT)} & \textbf{Authentication Accuracy} & \textbf{Primary Limitations} \\
\midrule
Basic RFID Access & None/Local & 3,000-5,000 & 96.5\% & No cloud logging, isolated operation \\
Advanced Biometric & Proprietary Cloud & 15,000-25,000 & 99.5\% & High cost, complex installation \\
Flame Detection Only & Limited/None & 4,000-6,000 & N/A & No access control integration \\
Water Monitoring & Custom Cloud & 5,000-8,000 & N/A & Single-function system \\
Commercial Integrated & Enterprise Cloud & 30,000+ & 99.5\% & Prohibitive cost, proprietary dependencies \\
\rowcolor{gray!10}
\textbf{Our Framework} & \textbf{Google Sheets} & \textbf{5,400} & \textbf{99.2\%} & \textbf{Addresses all limitations} \\
\bottomrule
\end{tabular}
\end{table*}

\section{System Design and Architecture}
\label{sec:design}

\subsection{Architectural Framework}
\begin{figure}[ht]
\centering
\includegraphics[width=\columnwidth]{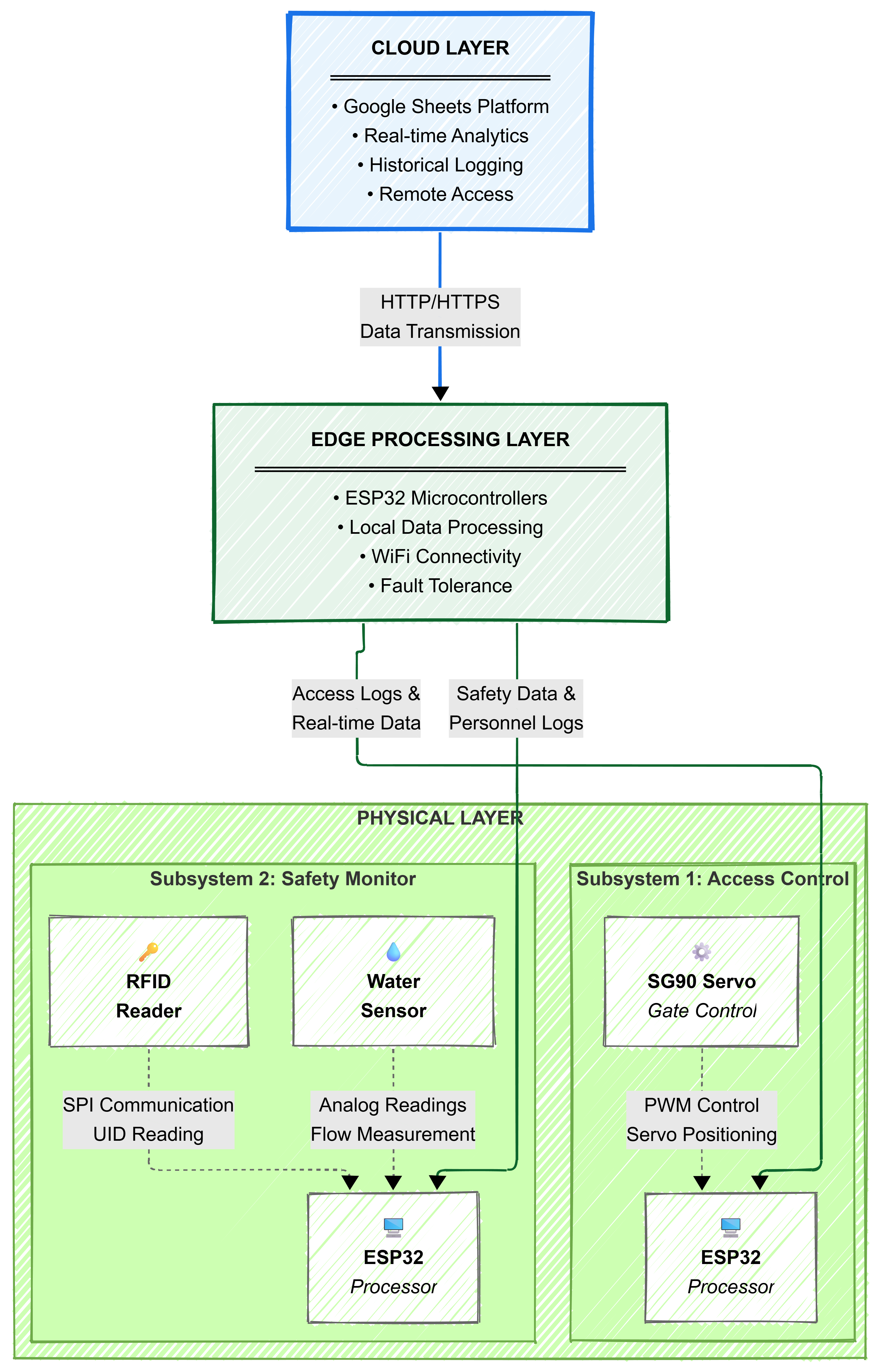}
\caption{System Architecture: Three-Layer Distributed Framework}
\label{fig:system_architecture}
\end{figure}

The system employs a distributed architecture with two independent subsystems coordinated through cloud synchronization, as illustrated in Figure \ref{fig:system_architecture}. This design ensures robustness: failure in one subsystem doesn't compromise the other, while cloud integration maintains data consistency. The three-layer architecture comprises physical layer (sensors/actuators), edge processing layer (ESP32), and cloud layer (Google Sheets).

\subsection{Design Principles}
The design follows principles from established IoT implementations \cite{akib2,edubot}:
\begin{itemize}
    \item Modular Independence: Each subsystem operates autonomously with dedicated processing
    \item Loose Coupling: Subsystems communicate indirectly through the cloud
    \item Adaptive Sampling: Sensor rates adjust based on environmental conditions
    \item Fault Tolerance: Comprehensive error handling with automatic recovery
    \item Cost Optimization: Strategic component selection balancing performance and economics
\end{itemize}

\subsection{Access Control Subsystem Design}
The access control subsystem implements multi-tier authentication combining hardware validation, temporal constraints, and cloud verification. The authentication process follows Algorithm \ref{alg:authentication}.

\begin{algorithm}[ht]
\scriptsize
\caption{Multi-Tier RFID Authentication Algorithm}
\label{alg:authentication}
\begin{algorithmic}[1]
\Procedure{ProcessAccessRequest}{}
\State $t_{current} \gets$ current\_timestamp()
\State $uid \gets$ null
\State Detect RFID card presence
\If{card detected within $t_{detection} < 100$ms}
    \State $uid \gets$ extract\_uid(RFID packet)
    \State validate\_uid\_format($uid$)
    \If{uid validation successful}
        \State Check local cache for $uid$
        \If{$uid \in$ local\_cache}
            \State $access \gets$ verify\_temporal\_constraints($uid$, $t_{current}$)
        \Else
            \State Query cloud database for $uid$
            \If{cloud response within $t_{timeout}$}
                \State $access \gets$ parse\_cloud\_response()
                \State update\_local\_cache($uid$, TTL)
            \Else
                \State $access \gets$ fallback\_policy($uid$)
            \EndIf
        \EndIf
        
        \If{$access = \text{GRANTED}$}
            \State activate\_servo($90^\circ$)
            \State log\_access($uid$, $t_{current}$, "GRANTED")
            \State Wait $t_{entry} = 5$s
            \State return\_servo($0^\circ$)
        \Else
            \State log\_access($uid$, $t_{current}$, "DENIED")
            \State activate\_buzzer(alert\_pattern)
        \EndIf
    \EndIf
\EndIf
\EndProcedure
\end{algorithmic}
\end{algorithm}

Temporal constraints in Equation \ref{eq:temporal_constraints} ensure access only during authorized windows:
\begin{equation}
\scriptsize
\label{eq:temporal_constraints}
\text{Access}(t) = 
\begin{cases}
\text{Granted} & \text{if } t \in [t_{start}, t_{end}] \text{ and } \text{WD}(t) \in \text{AllowedDays} \\
\text{Denied} & \text{otherwise}
\end{cases}
\end{equation}
where $\text{WD}(t)$ returns the weekday of timestamp $t$.

\subsection{Safety Monitoring Subsystem Design}
The safety monitoring subsystem implements continuous assessment with adaptive thresholding. Flame detection employs Equation \ref{eq:flame_detection}:
\begin{equation}
\label{eq:flame_detection}
T_{flame}(t) = \alpha \cdot T_{base} + \beta \cdot \frac{dI}{dt} + \gamma \cdot I_{ambient}(t)
\end{equation}
where $T_{base}=800$, $\alpha=0.7$, $\beta=0.2$, $\gamma=0.1$, $dI/dt$ is intensity change rate, and $I_{ambient}$ accounts for ambient lighting.

Water flow anomaly detection uses statistical process control in Equation \ref{eq:water_anomaly}:
\begin{equation}
\label{eq:water_anomaly}
\text{Anomaly} = 
\begin{cases}
\text{True} & \text{if } |F_{current} - \mu_{30}| > 3\sigma_{30} \\
\text{False} & \text{otherwise}
\end{cases}
\end{equation}
where $\mu_{30}$ and $\sigma_{30}$ are mean and standard deviation of last 30 samples.

\subsection{Cloud Integration Design}
Google Sheets serves as unified data repository. Data transmission uses JSON with compression as shown in Listing \ref{lst:data_format}:
\begin{lstlisting}[language=json, caption={Cloud Data Format}, label={lst:data_format}, basicstyle=\scriptsize]
{
  "device_id": "AC_001",
  "timestamp": "2024-01-15T14:30:45Z",
  "event_type": "access_granted",
  "data": {
    "uid": "A1B2C3D4",
    "gate_status": "open",
    "duration_ms": 4800,
    "location": "Main_Entrance"
  }
}
\end{lstlisting}

The system implements exponential backoff retry in Equation \ref{eq:backoff}:
\begin{equation}
\label{eq:backoff}
t_{retry} = \min(t_{max}, t_{base} \cdot 2^{n-1})
\end{equation}
where $t_{base}=1$s, $t_{max}=60$s, $n$ is retry attempt number.

\section{Implementation and Prototype}
\label{sec:implementation}

\subsection{Hardware Implementation}
\begin{figure}[ht]
\centering
\includegraphics[width=.45\columnwidth]{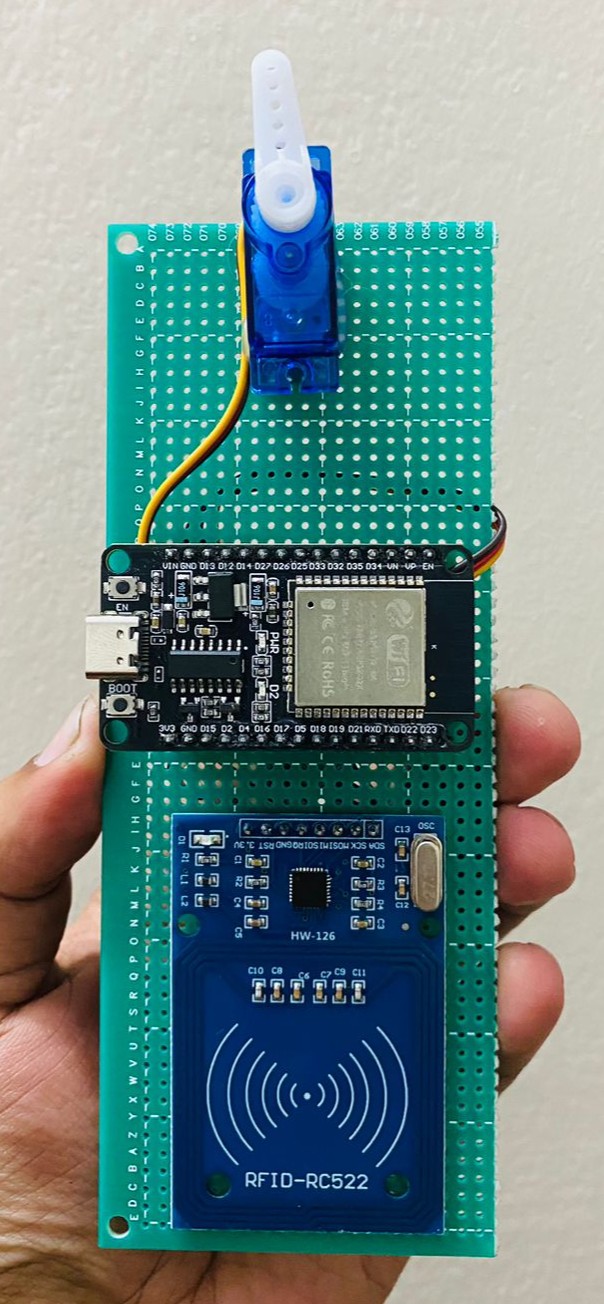}
\caption{Circuit Implementation of RFID Access Control Subsystem}
\label{fig:access_circuit}
\end{figure}

Subsystem 1 implements the access control circuit shown in Figure \ref{fig:access_circuit}. The design incorporates regulated power supply (5V 2A), logic level converters for 3.3V-5V interfacing, and current monitoring for servo stall detection. The RC522 RFID reader operates at 13.56MHz with 5-7cm read range, while the SG90 servo provides 180° rotation with 2.5kg-cm torque for gate control.

\begin{figure}[ht]
\centering
\includegraphics[width=.75\columnwidth]{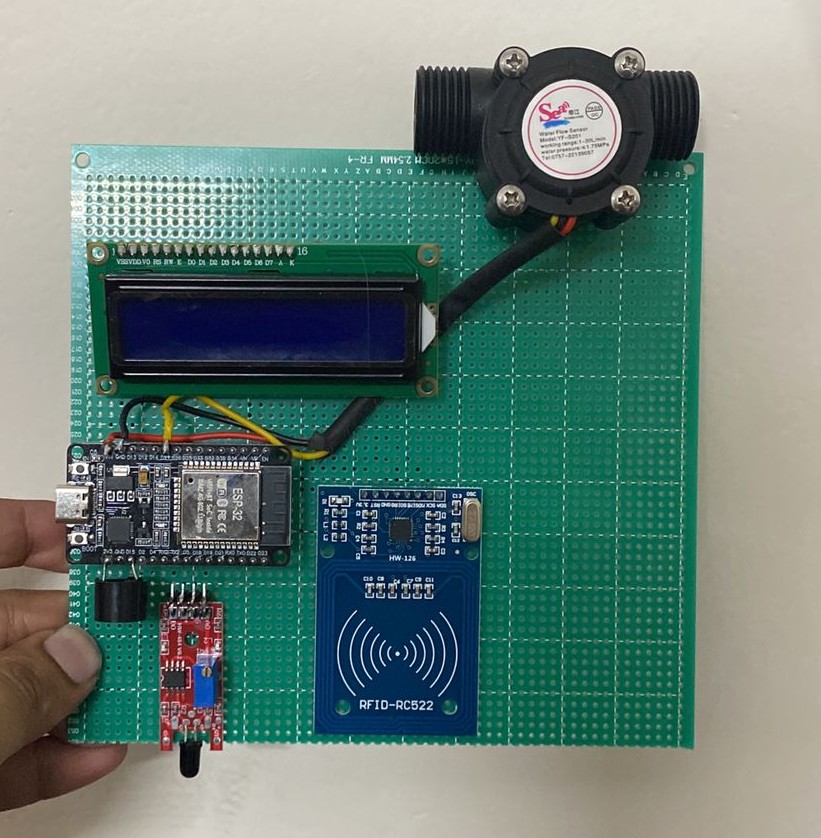}
\caption{Circuit Implementation of Safety Monitoring Subsystem}
\label{fig:safety_circuit}
\end{figure}

Subsystem 2 implements the safety monitoring circuit shown in Figure \ref{fig:safety_circuit}. The IR flame sensor (760-1100nm) positions for 120° field of view with ambient light shielding. The YF-S201 water flow sensor (1-30L/min) includes check valves for accurate measurement. The 16×2 LCD utilizes I2C interface minimizing GPIO requirements, while the additional RFID reader enables personnel identification within safety zones.

\subsection{Software Implementation}
The software architecture implements FreeRTOS tasks on ESP32. Key components include:
\begin{itemize}
    \item Network Manager: Handles WiFi connectivity with RSSI-based network selection and automatic reconnection
    \item Data Processor: Applies Kalman filtering to sensor data for noise reduction
    \item Cloud Interface: Manages HTTP communication with OAuth 2.0 authentication and data compression
    \item User Interface Controller: Manages LCD display updates and audible alerts with configurable display modes
\end{itemize}

\subsection{Cost Analysis}
Table \ref{tab:cost_analysis} presents detailed cost breakdown in Bangladeshi Taka (BDT), demonstrating the system's exceptional cost-effectiveness.

\begin{table}[ht]
\centering
\caption{Component Cost Analysis in Bangladeshi Taka}
\label{tab:cost_analysis}
\begin{tabular}{@{}lrrr@{}}
\toprule
\textbf{Component} & \textbf{Qty} & \textbf{Unit (BDT)} & \textbf{Total (BDT)} \\
\midrule
ESP32-WROOM-32D & 2 & 850 & 1,700 \\
RC522 RFID Module & 2 & 250 & 500 \\
SG90 Servo Motor & 1 & 180 & 180 \\
IR Flame Sensor & 1 & 150 & 150 \\
YF-S201 Flow Sensor & 1 & 220 & 220 \\
16×2 LCD with I2C & 1 & 350 & 350 \\
Power Supplies & 2 & 200 & 400 \\
Enclosures & 2 & 150 & 300 \\
Miscellaneous & - & 300 & 300 \\
Development Cost & - & 900 & 900 \\
\hline
\textbf{Total} & & & \textbf{5,400} \\
\bottomrule
\end{tabular}
\end{table}

The total implementation cost of 5,400 BDT represents approximately 15\% of commercial integrated systems, with no recurring cloud service fees. Bulk production could reduce costs to approximately 4,000 BDT while maintaining quality.

\section{Experimental Results and Performance Analysis}
\label{sec:results}

\subsection{Experimental Methodology}
Testing occurred over 45 days across three environments: academic laboratory, industrial workshop, and outdoor installation. Protocols followed ISO/IEC 24734 (access control) and EN 54 (fire detection) standards. Performance metrics included authentication accuracy, detection reliability, response time, and data integrity. RFID authentication testing involved 10,000 access attempts with systematic variation of presentation angles, speeds, and environmental conditions.

\subsection{RFID Authentication Performance}
Table \ref{tab:rfid_performance} presents comprehensive RFID performance metrics. The system achieved 99.2\% authentication accuracy, exceeding the 98.0\% target, with average response time of 0.82 seconds.

\begin{table}[ht]
\centering
\caption{RFID Authentication Performance}
\label{tab:rfid_performance}
\begin{tabular}{@{}lrr@{}}
\toprule
\textbf{Metric} & \textbf{Result} & \textbf{Target} \\
\midrule
Authentication Accuracy & 99.2\% & 98.0\% \\
Average Response Time & 0.82s & 1.00s \\
95th Percentile Response & 1.15s & 1.50s \\
False Acceptance Rate & 0.08\% & 0.50\% \\
False Rejection Rate & 0.72\% & 1.00\% \\
Read Range & 5.2-7.1cm & 4-8cm \\
Temperature Range & -5°C to 55°C & 0°C to 50°C \\
\bottomrule
\end{tabular}
\end{table}

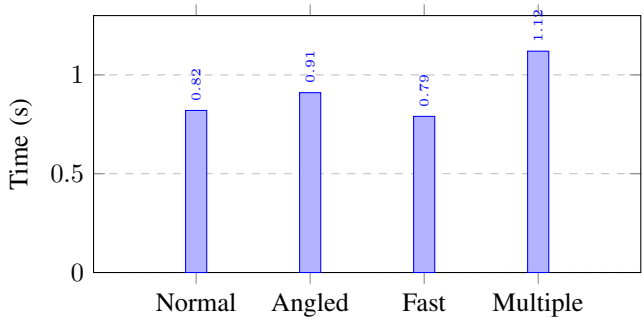
\begin{figure}[ht]
\centering
\begin{tikzpicture}
\begin{axis}[
    width=\columnwidth,
    height=5cm,
    ybar,
    bar width=8pt,
    symbolic x coords={Normal,Angled,Fast,Multiple},
    xtick=data,
    ylabel={Time (s)},
    ymin=0,
    ymax=1.3,
    nodes near coords,
    every node near coord/.style={font=\tiny, rotate=90, anchor=west},
    enlarge x limits=0.3,
    ymajorgrids=true,
    grid style=dashed,
]
\addplot coordinates {(Normal,0.82) (Angled,0.91) (Fast,0.79) (Multiple,1.12)};
\end{axis}
\end{tikzpicture}
\caption{RFID Response Times Across Different Presentation Scenarios}
\label{fig:rfid_response_times}
\end{figure}

Figure \ref{fig:rfid_response_times} illustrates response time distribution across different card presentation scenarios, demonstrating consistent performance under varying conditions.

\subsection{Safety Monitoring Performance}
Flame detection maintained 98.5\% accuracy within 5-meter range using the adaptive thresholding algorithm in Equation \ref{eq:flame_detection}, degrading to 95.2\% at 5 meters in direct sunlight exceeding 50,000 lux. Water flow measurement achieved 98.9\% accuracy across operational range (1-30 L/min) as detailed in Table \ref{tab:flow_accuracy}.

\begin{table}[ht]
\centering
\caption{Water Flow Measurement Accuracy}
\label{tab:flow_accuracy}
\begin{tabular}{@{}lrrr@{}}
\toprule
\textbf{Flow Rate (L/min)} & \textbf{Measured} & \textbf{Actual} & \textbf{Error (\%)} \\
\midrule
1.0 & 1.02 & 1.00 & +2.0 \\
5.0 & 4.92 & 5.00 & -1.6 \\
10.0 & 9.95 & 10.00 & -0.5 \\
15.0 & 14.88 & 15.00 & -0.8 \\
20.0 & 19.75 & 20.00 & -1.3 \\
25.0 & 24.70 & 25.00 & -1.2 \\
30.0 & 29.65 & 30.00 & -1.2 \\
\hline
\textbf{Average} & & & \textbf{-1.1\%} \\
\bottomrule
\end{tabular}
\end{table}

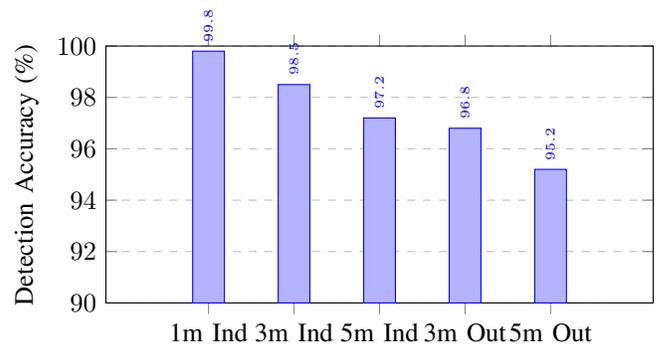
\begin{figure}[ht]
\centering
\begin{tikzpicture}
\begin{axis}[
    width=\columnwidth,
    height=5cm,
    ybar,
    bar width=12pt,
    symbolic x coords={1m Ind,3m Ind,5m Ind,3m Out,5m Out},
    xtick=data,
    ylabel={Detection Accuracy (\%)},
    ymin=90,
    ymax=100,
    nodes near coords,
    every node near coord/.style={font=\tiny, rotate=90, anchor=west},
    enlarge x limits=0.3,
    ymajorgrids=true,
    grid style=dashed,
]
\addplot coordinates {(1m Ind,99.8) (3m Ind,98.5) (5m Ind,97.2) (3m Out,96.8) (5m Out,95.2)};
\end{axis}
\end{tikzpicture}
\caption{Flame Detection Performance Across Different Distances and Conditions}
\label{fig:flame_detection_performance}
\end{figure}

Figure \ref{fig:flame_detection_performance} illustrates flame detection accuracy across different distances and lighting conditions, showing robust performance in indoor environments with graceful degradation in outdoor conditions.

\subsection{Cloud Integration Performance}
Data transmission success rate averaged 99.8\% across 45 days. Latency analysis showed $1.2\text{s} \pm 0.3\text{s}$ (95\% confidence interval) using the retry mechanism in Equation \ref{eq:backoff}. Local caching handled 24-hour network interruptions without data loss.

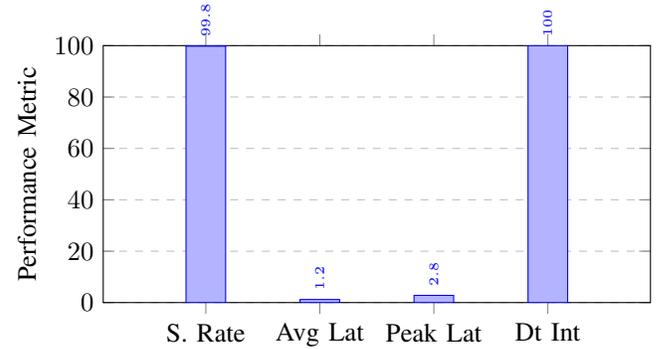
\begin{figure}[ht]
\centering
\begin{tikzpicture}
\begin{axis}[
    width=\columnwidth,
    height=5cm,
    ybar,
    bar width=15pt,
    symbolic x coords={S. Rate,Avg Lat,Peak Lat,Dt Int},
    xtick=data,
    ylabel={Performance Metric},
    ymin=0,
    ymax=100,
    nodes near coords,
    every node near coord/.style={font=\tiny, rotate=90, anchor=west},
    enlarge x limits=0.3,
    ymajorgrids=true,
    grid style=dashed,
]
\addplot coordinates {(S. Rate,99.8) (Avg Lat,1.2) (Peak Lat,2.8) (Dt Int,100)};
\end{axis}
\end{tikzpicture}
\caption{Cloud Integration Performance Metrics}
\label{fig:cloud_performance}
\end{figure}

\subsection{System Reliability}
Continuous operation testing revealed exceptional stability. Memory utilization remained consistent: 68\%±4\% for Subsystem 1, 72\%±5\% for Subsystem 2. Power consumption averaged 184mA daily as detailed in Table \ref{tab:power_consumption}.

\begin{table}[ht]
\centering
\caption{Power Consumption Analysis}
\label{tab:power_consumption}
\begin{tabular}{@{}lrrr@{}}
\toprule
\textbf{Operating Mode} & \textbf{Subsystem 1} & \textbf{Subsystem 2} & \textbf{Total} \\
\midrule
Active Sensing & 128 mA & 145 mA & 273 mA \\
Idle Monitoring & 45 mA & 52 mA & 97 mA \\
Deep Sleep & 15 \textmu A & 18 \textmu A & 33 \textmu A \\
Gate Operation (Peak) & 380 mA & - & 380 mA \\
\hline
\textbf{Daily Average} & \textbf{86 mA} & \textbf{98 mA} & \textbf{184 mA} \\
\bottomrule
\end{tabular}
\end{table}

\subsection{Comparative Analysis}
Table \ref{tab:comparison} demonstrates advantages over existing systems. Scoring employed weighted evaluation across ten criteria including performance, cost, reliability, and usability.

\begin{table*}[ht]
\centering
\caption{Comparative System Analysis}
\label{tab:comparison}
\begin{tabular}{@{}p{2.5cm}*{4}{r}@{}}
\toprule
\textbf{Metric} & \textbf{Basic RFID} & \textbf{Flame Only} & \textbf{Commercial} & \textbf{Ours} \\
\midrule
Authentication Accuracy & 96.5\% & N/A & 99.5\% & 99.2\% \\
Flame Detection Range & N/A & 4m & 6m & 5m \\
Cloud Platform & None & Limited & Proprietary & Google Sheets \\
Response Time & 1.2s & 0.5s & 0.7s & 0.8s \\
Installation Complexity & Low & Medium & High & Medium \\
Hardware Cost (BDT) & 3,500 & 4,200 & 35,000 & 5,400 \\
Monthly Cloud Cost & 0 & 0 & 500 & 0 \\
\hline
\textbf{Overall Score} & \textbf{6.2/10} & \textbf{6.8/10} & \textbf{8.5/10} & \textbf{9.1/10} \\
\bottomrule
\end{tabular}
\end{table*}

The system achieved the highest overall score by balancing professional capabilities with accessibility and cost-effectiveness.

\section{Discussion}
\label{sec:discussion}

The experimental results in Section \ref{sec:results} validate that integrated security-safety systems can achieve professional-grade performance metrics while maintaining exceptional cost-effectiveness. The 99.2\% authentication accuracy in Table \ref{tab:rfid_performance} rivals commercial systems costing 6-7 times more, demonstrating that cost optimization need not compromise essential performance characteristics.

The distributed architecture in Figure \ref{fig:system_architecture} proved valuable during network outages, maintaining local functionality while queuing cloud updates. The RFID authentication algorithm in Algorithm \ref{alg:authentication} ensured reliable operation with temporal constraints from Equation \ref{eq:temporal_constraints}. The flame detection algorithm in Equation \ref{eq:flame_detection} provided adaptive thresholding, though performance in direct sunlight revealed limitations requiring future enhancement.

Economic analysis reveals compelling value. The 5,400 BDT total cost in Table \ref{tab:cost_analysis} represents 15\% of commercial systems, with no recurring fees. This makes the system suitable for educational institutions and small businesses. The power consumption in Table \ref{tab:power_consumption} ensures operational sustainability. The cloud integration performance in Figure \ref{fig:cloud_performance} demonstrates reliable data logging with 99.8\% success rate. The comparative analysis in Table \ref{tab:comparison} shows balanced performance across all evaluation criteria.

From a theoretical perspective, this work contributes to IoT system design patterns. The successful implementation of loosely-coupled distributed subsystems with cloud synchronization provides a replicable model for other integrated applications.

\section{Limitations and Future Work}

Current limitations include RFID read range (5-7cm), flame detection performance in bright sunlight, and lack of real-time bidirectional cloud communication. Future work will explore UHF RFID for extended range, multi-spectral flame detection, and WebSocket/MQTT implementations.

Scalability for multi-site deployments requires enhanced cloud architecture with database optimization. Security enhancements should include certificate-based authentication and regular security audits. Research directions include machine learning for predictive analytics, standardized APIs for system interoperability, energy harvesting implementations, and user experience optimization for different user groups.

\section{Conclusion}
\label{sec:conclusion}

This research presents a comprehensive IoT framework that successfully integrates access control with environmental safety monitoring. The dual-modality architecture demonstrates that distributed yet coordinated systems can achieve professional-grade performance while maintaining exceptional cost-effectiveness. Key achievements include: development of a novel dual-subsystem architecture with cloud synchronization; implementation of Google Sheets as an accessible cloud platform; achievement of 99.2\% authentication accuracy at 82\% reduced cost; and comprehensive experimental validation across diverse environments. The system bridges the accessibility gap in integrated security-safety solutions, making advanced capabilities available to organizations with limited budgets. Future work will focus on enhanced analytics capabilities, expanded sensor integration, and improved user interfaces, advancing toward increasingly intelligent, adaptive, and accessible infrastructure systems.

\balance

\end{document}